\documentclass[lettersize,journal]{IEEEtran}
\usepackage{amsmath,amsfonts}
\usepackage{algorithmic}
\usepackage{algorithm}
\usepackage{array}
\usepackage[caption=false,font=normalsize,labelfont=sf,textfont=sf]{subfig}
\usepackage{textcomp}
\usepackage{stfloats}
\usepackage{url}
\usepackage{verbatim}
\usepackage{graphicx}
\usepackage{cite}
\usepackage{hyperref}
\usepackage{booktabs}
\usepackage{graphicx}
\usepackage{caption}
\usepackage{multirow}
\usepackage{color}
\begin{document}

\title{HDMba: Hyperspectral Remote Sensing Imagery Dehazing with State Space Model}

\author{IEEE Publication Technology,~\IEEEmembership{Staff,~IEEE,}
\thanks{This paper was produced by the IEEE Publication Technology Group. They are in Piscataway, NJ.}
\thanks{Manuscript received April 19, 2021; revised August 16, 2021.}}

\author{Hang Fu, Genyun Sun,~\IEEEmembership{Senior Member,~IEEE}, Yinhe Li, Jinchang Ren,~\IEEEmembership{Senior Member,~IEEE}, Aizhu Zhang,~\IEEEmembership{Member,~IEEE}, Cheng Jing, Pedram Ghamisi,~\IEEEmembership{Senior Member,~IEEE}
\thanks{H. Fu, G. Sun, A. Zhang and C. Jing are with the College of Oceanography and Space Informatics, China University of Petroleum (East China), Qingdao 266580, China.(e-mail: hangf\_upc@163.com; genyunsun@163.com) (\textit{Corresponding Author: Genyun Sun})}
\thanks{Y. Li and J. Ren are with the National Subsea Centre, Robert Gordon University, Aberdeen AB10 7AQ, U.K.(e-mail: jinchang.ren@ieee.org)}
\thanks{P. Ghamisi is with Helmholtz Institute Freiberg for Resource Technology, Helmholtz-Zentrum Dresden-Rossendorf, 09599 Freiberg, Germany.(e-mail: p.ghamisi@hzdr.de)}}

\markboth{Journal of \LaTeX\ 2024}%
{Shell \MakeLowercase{\textit{et al.}}: A Sample Article Using IEEEtran.cls for IEEE Journals}

\maketitle

\begin{abstract}
Haze contamination in hyperspectral remote sensing images (HSI) can lead to spatial visibility degradation and spectral distortion. Haze in HSI exhibits spatial irregularity and inhomogeneous spectral distribution, with few dehazing networks available. Current CNN and Transformer-based dehazing methods fail to balance global scene recovery, local detail retention, and computational efficiency. Inspired by the ability of Mamba to model long-range dependencies with linear complexity, we explore its potential for HSI dehazing and propose the first HSI Dehazing Mamba (HDMba) network. Specifically, we design a novel window selective scan module (WSSM) that captures local dependencies within windows and global correlations between windows by partitioning them. This approach improves the ability of conventional Mamba in local feature extraction. By modeling the local and global spectral-spatial information flow, we achieve a comprehensive analysis of hazy regions. The DehazeMamba layer (DML), constructed by WSSM, and residual DehazeMamba (RDM) blocks, composed of DMLs, are the core components of the HDMba framework. These components effectively characterize the complex distribution of haze in HSIs, aiding in scene reconstruction and dehazing. Experimental results on the Gaofen-5 HSI dataset demonstrate that HDMba outperforms other state-of-the-art methods in dehazing performance. The code will be available at {\color{blue}https://github.com/RsAI-lab/HDMba}.
\end{abstract}


\section{Introduction}
\IEEEPARstart{O}{ptical} remote sensing (RS) hyperspectral imagery (HSI) captures hundreds of contiguous narrow spectral bands, enabling the detection of subtle variations in ground surface characteristics. This capability makes HSI indispensable in environmental monitoring, agricultural assessment, and urban management~\cite{0,1}. However, atmospheric disturbances, such as haze, can cause significant distortions in spectral signatures, compromising the accuracy of surface feature identification and classification. Consequently, effective haze removal is essential to preserve the integrity and advantages of HSI in diverse remote sensing applications \cite{9618664}.

Conventional RS dehazing methods primarily rely on atmospheric scattering models and dark-object subtraction (DOS) models. Classical methods, such as the dark channel prior \cite{2} and haze thickness map \cite{3}, are commonly used for model parameter estimation. Kang et al. \cite{4} developed a DOS model-based HSI defogging model (Defog). These methods are often limited in dehazing efficacy and generalization due to their dependence on parameter settings and manual intervention. Deep learning techniques, particularly convolutional neural networks (CNNs), have been applied to RS dehazing, with models like FFANet \cite{5}, RSDehazeNet \cite{6}, CANet \cite{7}, and LKDNet \cite{8} demonstrating notable generalization performance by mapping hazy images directly to clear images. Recent studies have integrated attention mechanisms in HSI dehazing, resulting in models like SGNet \cite{9} and AACNet \cite{10}. Despite these advances, the limited receptive field of convolutional kernels and pixel-to-pixel translation challenges these methods in capturing long-range contextual information in hazy regions, leading to scene structure discrepancies.

\begin{figure}[!t]
\centering
\includegraphics[width=3.0in]{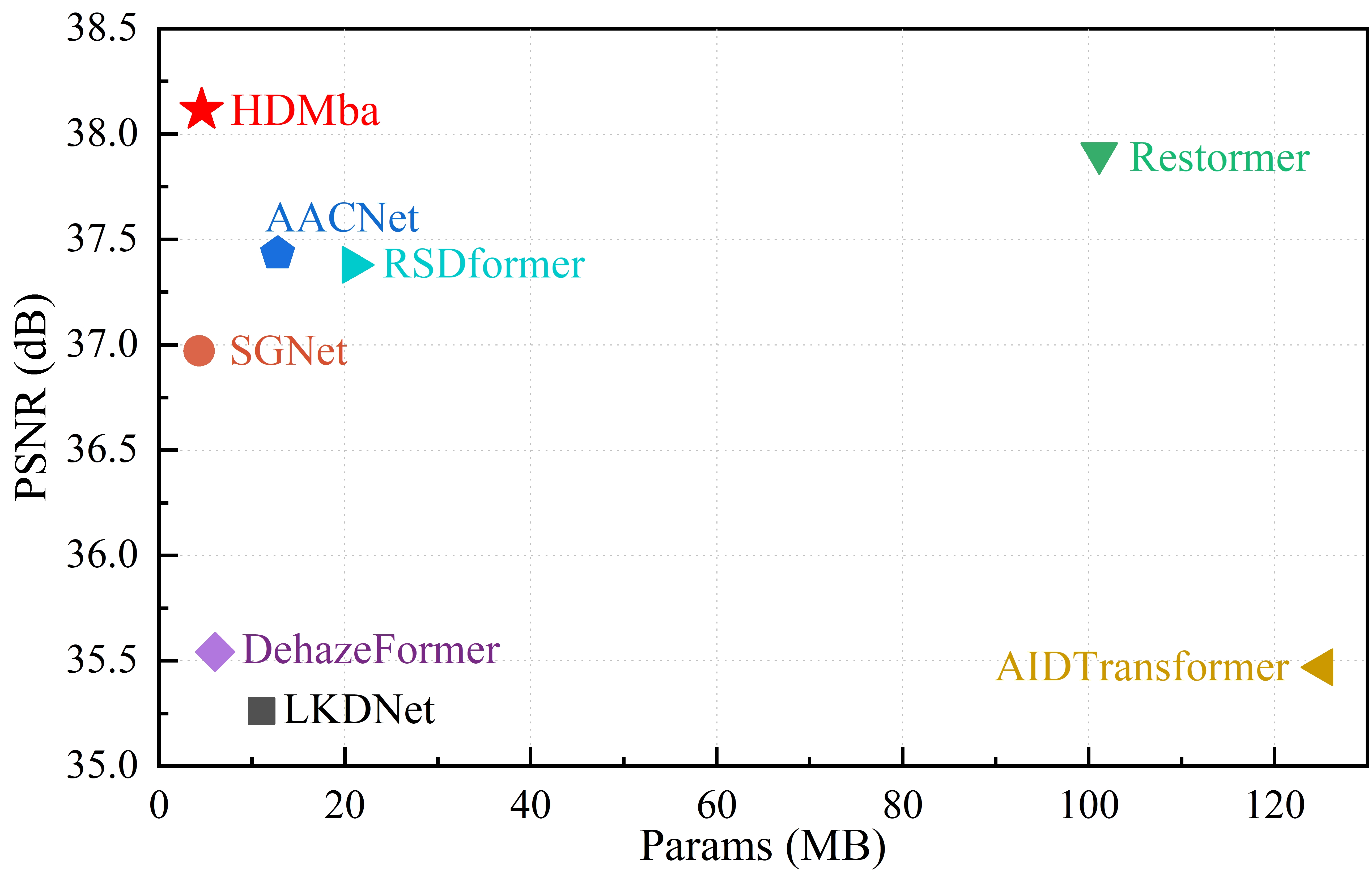}
\caption{Comparison of dehazing complexity and performance characterized by PSNR and parameters between our HDMba and other state-of-the-art networks on the Gaofen-5 Hyperspectral Dataset. Parameter calculation is based on image size of 64×64×305.}
\label{fig_1}
\end{figure}

Thanks to its superior global modelling capability, the Transformer has been successfully applied to remote sensing dehazing, yielding impressive results. Models such as Restormer \cite{11}, DehazeFormer \cite{12}, and AIDTransformer \cite{13} used U-shaped structures to extract deep global structural information. RSDformer \cite{14} further enhanced image structure recovery by incorporating novel self-attention mechanisms to capture both local and global correlations. However, the quadratic complexity of self-attention and the number of tokens leads to significant computational overhead on image dehazing. Recently, Mamba \cite{15}, a novel state space model (SSM) has shown great potential in long-sequence modelling with linear complexity. It has been applied to various RS tasks, including semantic segmentation, pan-sharpening, and denoising \cite{16}. While Mamba has been attempted for natural and RS-RGB image dehazing \cite{20,21}, its efficacy in HSI dehazing remains unexplored.

Compared to conventional image dehazing, haze in HSI exhibits irregular and locally significant inhomogeneities in the spatial domain, with shortwave bands being more sensitive to haze than longwave bands. We aim to leverage Mamba to explore the complex haze distribution in HSI, achieving efficient haze removal and scene reconstruction. To this end, we developed the first {\bf{H}}SI {\bf{D}}ehazing {\bf{M}}amba (HDMba) framework and designed a window selective scan module (WSSM) to model the local-global spectral-spatial information flow, effectively reconstructing the scene and spectral details in HSI. The main contributions of this work are summarized as follows:

1) We introduce HDMba, the first framework to explore the potential of Mamba for HSI dehazing. The designed DehazeMamba block integrates SSM, convolution, and residual learning to effectively model complex haze distributions in HSI, recovering scene structure and texture details.

2) We propose a new WSSM that captures local dependencies along with contextual global interactions, improving the perception of local haze regions and the characterization of differences between hazy and haze-free regions.

3) We construct an HSI dehazing dataset with 2,000 image pairs using the Gaofen-5 Advanced Hyperspectral Imager (AHSI) sensor. This dataset, along with the available hyperspectral defogging dataset (HDD), is used to assess the dehazing performance and complexity of HDMba against other state-of-the-art dehazing networks.

\begin{figure*}[!t]
\centering
\includegraphics[width=5.0in]{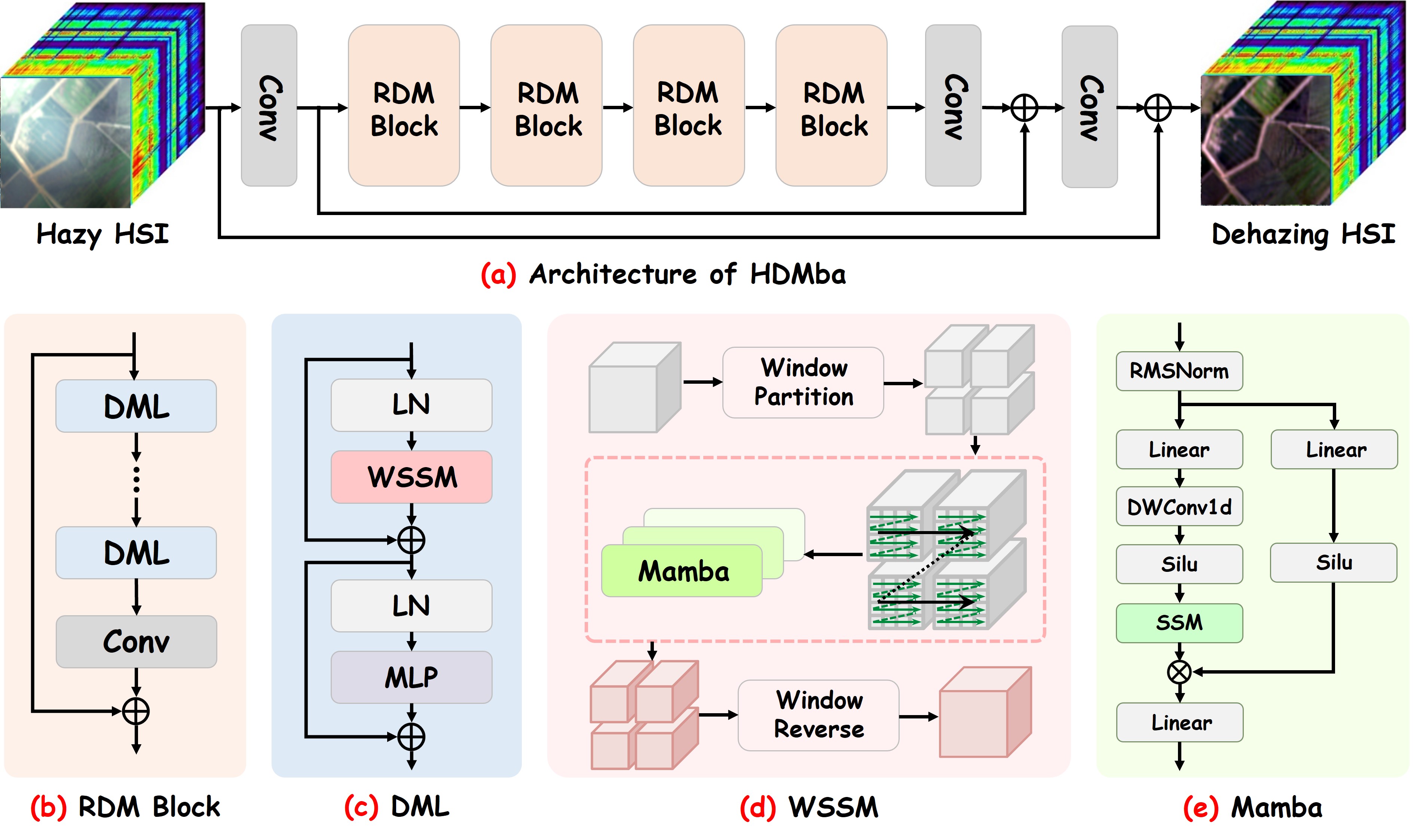}
\caption{(a) Overall architecture of HDMba. HDMba consists of multiple residual DehazeMamba (RDM) blocks and convolutional layers for end-to-end image dehazing. (b) RDM block comprises multiple DehazeMamba layers (DML) and a convolutional layer. (c) DML comprises Window selective scan module (WSSM) and MLP. (d) WSSM consists of window partition, (e) Mamba, and window reverse.}
\label{fig_2}
\end{figure*}

\section{Proposed method}
In this section, we first present the overall network framework of the proposed HDMba. Next, we introduce the key module, the Residual DehazeMamba (RDM) block, which consists of multiple DehazeMamba layers (DML). Finally, we describe the WSSM within the DML, detailing its approach to modelling and processing local and global information flows using Mamba.

\subsection{Network Architecture}
To ensure effective and direct dehazing feature extraction, we adopted an end-to-end multi-scale feature extraction framework based on RDM blocks, as shown in Fig. \ref{fig_2} (a). Downsampling in the U-net is discarded, which allows the high-frequency information to be preserved. HDMba first applies a 3×3 convolution layer to extract low-level features \({F}_{0}\in \mathbb{R}^{W\times H\times C}\) from the haze image \({X}\in \mathbb{R}^{W\times H\times B}\), where \(W\) and \(H\) represent the spatial dimensions, and \(C\) and \(B\) are the number of feature channels and the number of input bands, respectively. Then, \({F}_{0}\) is processed through several RDM blocks to extract deep features for scene recovery with the feature size of \(W\times H\times C\), and this process can be expressed as: 
\begin{equation}
\mathcal{F}_{i} =\mathrm{RDM} _{i} (\mathcal{F}_{i-1}),i=1,2,...,I
\end{equation}
where \(\mathrm{RDM} _{i}\) represents the \(i\)-th RDM block, and \(\mathcal{F}_{i}\) denotes the \(i\)-th spatial-spectral feature obtained by this block. The network is designed with 4 RDM blocks. To recover a clean scene \(\mathcal{Y}\in \mathbb{R}^{W\times H\times B}\) from the deep feature \(\mathcal{F}_{I}\), two 3×3 convolution layers are concatenated with the shallow feature via skip connections. The global skip connection between the hazy image and the output enables the intermediate network to learn the irregular and uneven haze characteristics distributed across the spectral and spatial domains. A combination of mean squared error (MSE) and \(L_{1} \) norm is used as loss function for network training:
\begin{equation}
\mathcal{L}=\theta _{1}  \left \| \mathcal{Y} -\mathcal{X}  \right \| _{mse} +\theta _{2}\left \|\mathcal{Y} -\mathcal{X}  \right \|_{1}   \end{equation}
where \(\left \|\cdot   \right \|_{mse}   \) represents the MSE loss and \(\left \|\cdot   \right \|_{1}   \) represents the \(L_{1} \) norm. \(\theta _{1}\) and \(\theta _{2}\) are the weighting factors.

\subsection{Residual DehazeMamba block}
Haze exhibits an irregular spatial distribution, making it essential to focus on local haze areas and extract crucial information from them for scene recovery. We designed the RDM block (Fig. \ref{fig_2} (b)) for deep local and long-range information modelling, which mainly contains multiple DMLs, represented as follows:
\begin{equation}
\mathcal{F}_{k} =\mathrm{DML} _{k} (\mathcal{F}_{k-1}), k=1,2,...,K
\end{equation}
where \(\mathrm{DML} _{k}\) represents the \(k\)-th DML and \(\mathcal{F}_{k}\) denotes the \(k\)-th spatial-spectral feature in RDM block. Each RDM block concludes with a 3×3 convolution layer, adding the residual features from the previous block via a skip connection. For each DML (Fig. \ref{fig_2} (c)), we combine the normalized layer with WSSM and MLP respectively to improve the nonlinearity characterization while modelling spatial information. The process can be defined as follows:
\begin{equation}
\mathcal{F}_{t}' =\mathrm{WSSM} (\mathrm{LN} (\mathcal{F}_{t-1}))+\mathcal{F}_{t-1}
\end{equation}
\begin{equation}
\mathcal{F}_{t} =\mathrm{MLP} (\mathrm{LN} (\mathcal{F}_{t}'))+\mathcal{F}_{t}'
\end{equation}
where \(\mathcal{F}_{t-1}\) represents the input feature embedding of DML, \(\mathcal{F}_{t}'\) and \(\mathcal{F}_{t}\) represent the output of \(\mathrm{WSSM}\) and \(\mathrm{MLP}\), respectively, and \(\mathrm{LN}\) represents the Layer Normalization layer.

\subsection{Window selective scan module}
Currently, most visual Mamba approaches primarily capture long-term dependencies by increasing scanning directions, but they lack the ability to effectively capture local spatial details and inter-regional correlations \cite{19}. We proposed a novel WSSM to enhance the processing capacity in local areas where haze is distributed, as shown in Fig. \ref{fig_2} (d).

Specifically, for the input feature embedding \(\mathcal{Z} _{ip} \in \mathbb{R}^{W\times H\times C} \), the window partition \cite{18} is first performed in spatial dimension with the window size of \(M\), resulting in \(\frac{W\times H}{M^{2} } \) overlapping patches \(z_{ip} \in \mathbb{R}^{M^{2} \times C} \). These patches then capture local dependencies through Mamba, while retaining the correlation of different local regions, ensuring a comprehensive analysis of haze regions in the image. Finally, through a window reverse operation, the output patches \(z_{op} \in \mathbb{R}^{M^{2} \times C} \) from Mamba are gathered to obtain the final features \(\mathcal{Z} _{op} \in \mathbb{R}^{W\times H\times C} \). This process can be expressed as follows:
\begin{equation}
\left \{  z_{ip}\right \} =\mathrm{WinPartition} (\mathcal{Z} _{ip})
\end{equation}
\begin{equation}
\left \{  z_{op}\right \} = \mathrm{Mamba} (\left \{  z_{ip}\right \}) 
\end{equation}
\begin{equation}
\mathcal{Z} _{op} = \mathrm{WinReverse} (\left \{  z_{op}\right \})
\end{equation}
where \(\left \{z _{ip}  \right \} \in \mathbb{R}^{\frac{WH}{M^{2} }\times   M^{2} \times C} \) and \(\left \{z _{op}  \right \} \in \mathbb{R}^{\frac{WH}{M^{2} }\times   M^{2} \times C} \). For Mamba (Fig. \ref{fig_2} (e)), the input spatial feature sequence enters two branches after passing through RMSNorm. In the main branch, the features undergo successive processing by a linear layer, depth-wise separable convolution, a SiLU activation function, and SSM, effectively integrating haze characteristics from various regions. The other branch passes through a linear layer and a SiLU function and multiplies the output of the main branch. A normalization layer produces the final output. The process can be represented as follows:
\begin{equation}
z_{op1} =\mathrm{SSM} (\mathrm{\phi} (\mathrm{DConv} (\mathrm{linear} (\mathrm{RMSNorm}(z_{ip})))))
\end{equation}
\begin{equation}
z_{op2} =\mathrm{\phi} (\mathrm{linear} (\mathrm{RMSNorm}(z_{ip})))
\end{equation}
\begin{equation}
z_{op} = \mathrm{linear} (z_{op1} \otimes z_{op2} )
\end{equation}
where \(z_{op1}\) and \(z_{op2}\) represent the output of the two branches respectively. \(\mathrm{DConv}\left ( \cdot  \right ) \) represents the depth-wise separable convolution, \(\mathrm{\phi }\left ( \cdot  \right )\) denotes the SiLU activation function and \(\otimes \) indicates the element-wise multiplication.

\section{Experimental results}
\subsection{Dataset and Implementation Details}
To evaluate the effectiveness of the proposed HDMba, we used two Gaofen-5 HSI datasets: HyperDehazing dataset\footnote{The dataset will be publicly available at https://github.com/RsAI-lab/HyperDehazing} and HDD \cite{4}. The HyperDehazing dataset is synthesized based on the DOS model using 100 clean scenes with 20 different haze thicknesses and 5 haze abundances. It contains a total of 2000 hazy and haze-free image pairs, each with a size of 512×512×305. 90\% of this dataset was used for network training, while the remaining 10\% was reserved for testing. HDD comprises 20 reference-free hazy images, each with dimensions 512×512×305. All images in this dataset were used for network testing. To meet memory requirements, we cropped the training data to 64×64 and the test data to 128×128. 

We set the number of DMLs \(K\)=4, the window size \(M\)=8. \(\theta _{1}\) and \(\theta _{2}\) are set to 1 and 0.1, respectively. The batch size is set to 4, and all datasets are trained for 10,000 iterations. The Adam optimization operator was employed to accelerate the training, where the momentum parameters were set to 0.9, 0.999, and 10\(^{-8}\), respectively. The initial learning rate was set to 1×10\(^{-4}\), with the cosine annealing strategy to adjust the learning rate. The whole network was implemented on the PyTorch framework with an NVIDIA GeForce RTX 3060 GPU.

\subsection{Dehazing results}
To quantitatively evaluate the dehazing performance, we used structural similarity index measurement (SSIM), peak signal-to-noise ratio (PSNR), universal image quality index (UQI), and spectral angle mapping (SAM) metrics for paired images, and natural image quality evaluator (NIQE) and average gradient (AG) metrics for real images. The results are shown in Table \ref{table1}. Transformer-based methods generally outperform CNN-based methods. However, HDMba achieves the best results across all metrics except UQI. When processing high-dimensional data, HDMba has significantly fewer parameters (4.60M) compared to most dehazing networks.

\begin{figure*}[!t]
\centering
\includegraphics[width=6.5in]{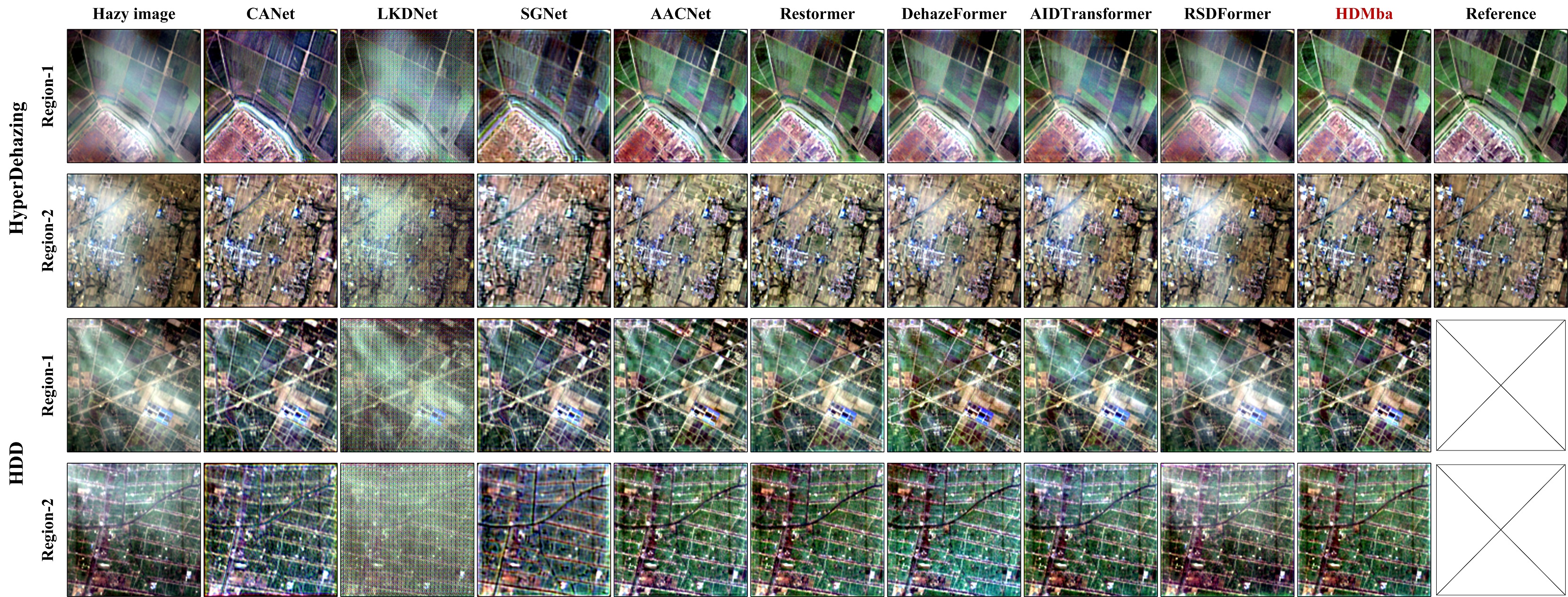}
\caption{Comparison of visual results on HyperDehazing and HDD.}
\label{fig_3}
\end{figure*}

To visualize the effectiveness of dehazing, Fig. \ref{fig_3} presents dehazing images of several state-of-the-art methods. It is evident that LKDNet, AIDTransformer, and RSDFormer do not completely remove the haze. The scenes recovered by CANet and SGNet exhibit spectral distortions. While AACNet and Restormer manage to recover parts of the clean scene, some haze residue remains. In contrast, the proposed HDMba recovers the result closest to the clean scene, with a good reconstruction of surface details.


\begin{table*}[ht]
\renewcommand{\arraystretch}{1.1}
\centering
\caption{Comparison of quantitative results on HyperDehazing and HDD. \textbf{Bold} and \underline{underlined} indicate best and second-best results \label{table1}}
\scalebox{1.1}{%
\begin{tabular}{ll|cccc|cc|c}
\hline
\multicolumn{2}{c|}{\textbf{Dataset}} & \multicolumn{4}{c|}{\textbf{HyperDehazing}} & \multicolumn{2}{c|}{\textbf{HDD}} & \textbf{Complexity} \\
\multicolumn{2}{c|}{\textbf{Methods}} & \textbf{SSIM$\uparrow$} & \textbf{PSNR$\uparrow$} & \textbf{UQI$\uparrow$} & \textbf{SAM$\downarrow$} & \textbf{NIQE$\downarrow$} & \textbf{AG$\uparrow$} & \textbf{Params (M)} \\
\hline
\multirow{1}{*}{Model-based} 
& Defog\cite{4} & 0.7020 & 27.5621 & 0.7853 & 0.2637 & 17.8467 & 0.1983 & - \\
\hline
\multirow{6}{*}{CNN-based} 
& FFANet\cite{5} & 0.9035 & 32.1437 & 0.9476 & 0.0891 & 18.6571 & 0.1357 & 4.69 \\
& RSDehazeNet\cite{6} & 0.9409 & 34.4354 & 0.9721 & 0.0672 & 17.9746 & 0.1470 & \textbf{1.51} \\
& CANet\cite{7} & 0.9542 & 34.6147 & 0.9743 & 0.0631 & 15.7951 & 0.2524 & 12.12 \\
& LKDMNet\cite{8} & 0.9448 & 35.2614 & 0.9618 & 0.0588 & 19.7972 & \underline{0.2616} & 11.06 \\
& SGNet\cite{9} & 0.9672 & 36.9704 & \underline{0.9785} & 0.0568 & 16.3662 & 0.2032 & \underline{4.32} \\
& AACNet\cite{10} & \underline{0.9734} & 37.4322 & \textbf{0.9797} & 0.0425 & 15.2740 & 0.2556 & 12.76 \\
\hline
\multirow{4}{*}{Transformer-based} 
& Restormer\cite{11} & 0.9702 & \underline{37.9080} & 0.9755 & 0.0421 & 15.7965 & 0.2396 & 101.16 \\
& DehazeFormer\cite{12} & 0.9708 & 35.5426 & 0.9739 & 0.0432 & \underline{14.3863} & 0.2456 & 6.04 \\
& AIDTransformer\cite{13} & 0.9723 & 35.4695 & 0.9736 & \underline{0.0401} & 15.6097 & 0.2330 & 125.11 \\
& RSDformer\cite{14} & 0.9709 & 37.3785 & 0.9743 & 0.0462 & 16.4790 & 0.2461 & 20.89 \\
\hline
\multirow{1}{*}{Mamba-based} 
& HDMba & \textbf{0.9763} & \textbf{38.1340} & 0.9765 & \textbf{0.0382} & \textbf{13.7959} & \textbf{0.2663} & 4.60 \\
\hline
\end{tabular}%
}
\end{table*}

\subsection{Spectrum reconstruction analysis}
\begin{figure}[!t]
\centering
\includegraphics[width=3.2in]{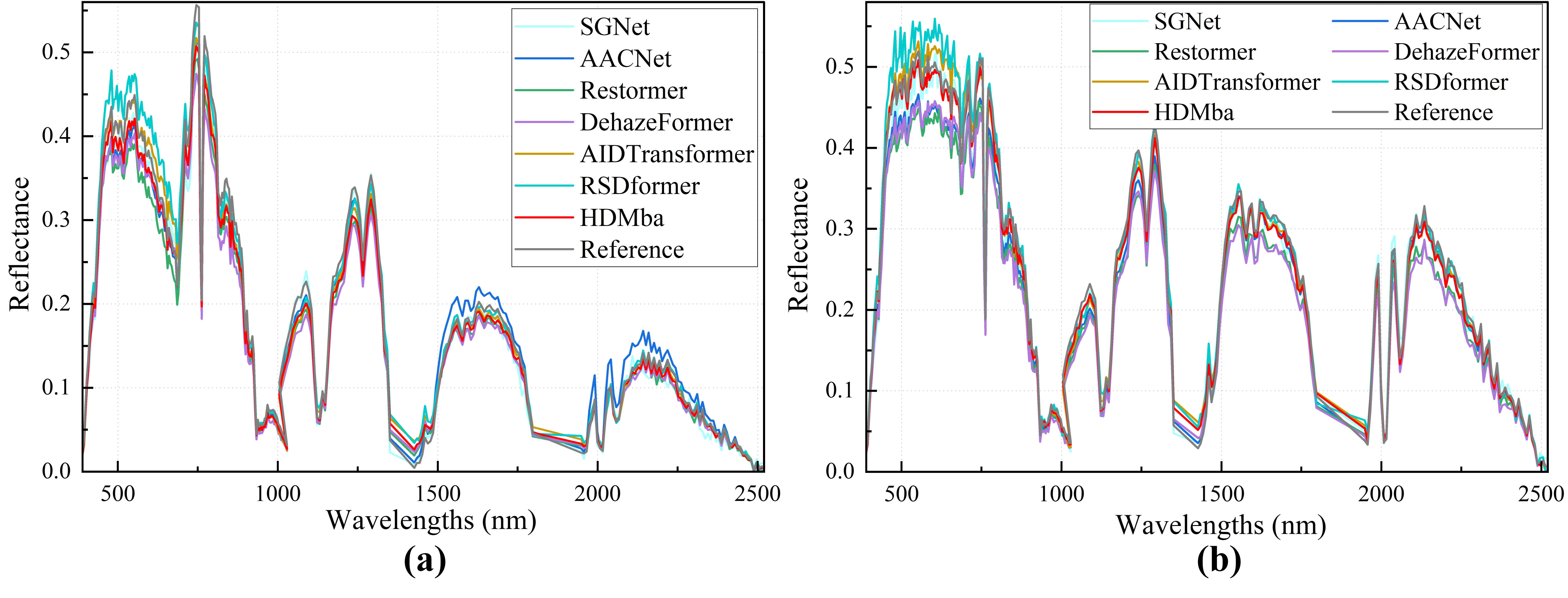}
\caption{Comparison of spectrum reconstruction performance from hazy HSIs on HyperDehazing. (a) Building scene. (b) Vegetable scene.}
\label{fig_4}
\end{figure}

As shown in Fig. \ref{fig_4}, we selected building and vegetation scenes to compare spectra reconstruction performance. RSDFormer has an insufficient dehazing ability, resulting in spectral curves significantly higher than the reference value. AIDTransformer exhibits similar issues (Fig. \ref{fig_4}(b)). AACNet, DehazeFormer, and Restormer produce spectra in the visible range that are lower than the reference, indicating excessive dehazing. In contrast, HDMba achieves spectra closest to the reference (Fig. \ref{fig_4} (b)), with spectral trends that are highly consistent despite some deviations in certain cases (Fig. \ref{fig_4} (a)).

\begin{figure}[!t]
\centering
\includegraphics[width=3.2in]{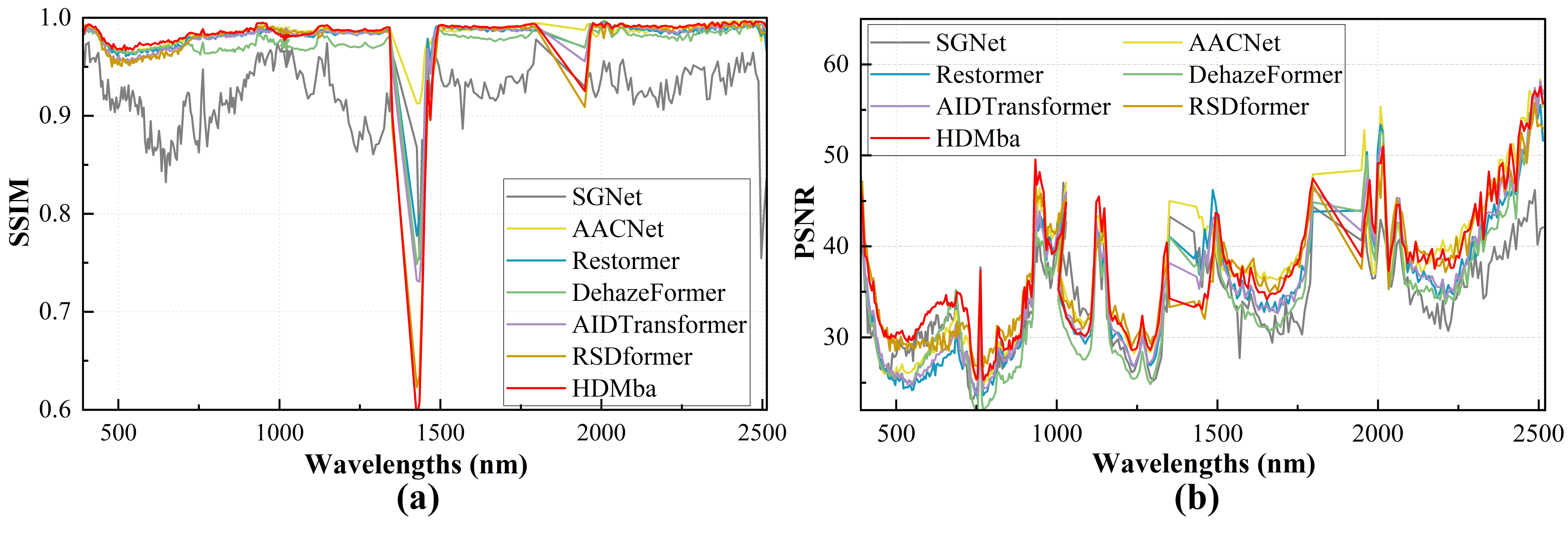}
\caption{Performance trends of various dehazing methods across wavelengths, measured by (a) SSIM and (b) PSNR.}
\label{fig_5}
\end{figure}

\subsection{Performance across wavelengths}
The effectiveness of HDMba in processing hazy HSIs across different bands is quantitatively evaluated using the SSIM and PSNR metrics, as shown in Fig. \ref{fig_5}. The results indicate that HDMba consistently outperforms other methods across most bands, achieving better performance than SGNet (which has the lowest SSIM) and DehazeFormer (which has the lowest PSNR). It should be noted that the performance of HDMba degrades in the 1430 nm and 1950 nm wavelength ranges, which are close to the water and atmospheric absorption bands and can cause severe interference. Nevertheless, our method exhibits excellent dehazing effects across nearly all spectral bands.

\subsection{Ablation Study}
We conducted ablation experiments on HyperDehazing to investigate the effectiveness of each component of the proposed model. These experiments included evaluating the impact of constituent elements within the DML, as well as exploring the effect of different partition window sizes in the WSSM on the dehazing results.

1) \emph{Analysis of DML}: We trained the network with variations in constituent elements of Mamba and MLP, presenting the corresponding dehazing results in Table \ref{table2}. It is evident that the combination of SSM, DWConv1d, and multiplication in Mamba yields excellent dehazing performance, with further enhancement achieved by incorporating MLP.

2) \emph{Analysis of window size}: We assessed the effect of window size in the WSSM on dehazing, and the results are presented in Table \ref{table3}. As we can see, larger window sizes lead to improved performance but also increase computational costs. We set the window size to 8 to strike a balance between performance and computation cost.

\begin{table}[ht]
\centering
\caption{Ablation analysis of constituent elements within the proposed DML\label{table2}}
\scalebox{1.0}{%
\begin{tabular}{ccccccc}
\toprule
\multicolumn{3}{c}{\textbf{Mamba}} & \multicolumn{1}{c}{\multirow{2.5}{*}{\textbf{MLP}}} & \multirow{2.5}{*}{\textbf{SSIM$\uparrow$}} & \multirow{2.5}{*}{\textbf{PSNR$\uparrow$}} & \multirow{2.5}{*}{\textbf{Params (M)}} \\
\cmidrule{1-3}
\textbf{SSM} & \textbf{DConv} & $\otimes$ &  &  &  &  \\
\hline
$\times$ & $\times$ & $\times$ & $\checkmark$ &  0.8062 & 28.1639 & 1.11 \\
\hline
$\checkmark$ & $\times$ & $ \times$ & $\times$ &  0.9696 & 36.1114 & 3.92 \\
\hline
$\checkmark$ & $\checkmark$ & $\times$ & $\times$ &  0.9680 & 36.2026 & 3.96 \\
\hline
$\checkmark$ & $\checkmark$ & $\checkmark$ & $\times$ & 0.9737 & 36.4216 & 4.35 \\
\hline
$\checkmark$ & $\checkmark$ & $\checkmark$ & $\checkmark$ & 0.9783 & 37.2427 & 4.60 \\
\hline
\end{tabular}%
}
\end{table}

\begin{table}[ht]
\centering
\caption{Analysis on the effect of window size in WSSM \label{table3}}
\scalebox{1.0}{%
\begin{tabular}{cccc}
\toprule
\textbf{Window size} & \textbf{SSIM$\uparrow$} & \textbf{PSNR$\uparrow$} & \textbf{Params (M)} \\
\hline
2 & 0.9733 &  37.7522 & 4.56 \\
\hline
4 & 0.9740 &  37.8846 & 4.57 \\
\hline
8 & 0.9763 &  38.1340 & 4.60 \\
\hline
16 & 0.9787 &  38.3088 & 4.74 \\
\hline
\end{tabular}%
}
\end{table}

\section{Conclusion}
In this paper, we propose a hyperspectral image dehazing network (HDMba) based on Mamba. The proposed residual DehazeMamba (RDM) blocks effectively characterize the complex haze distribution in HSI data, enhancing scene and texture detail recovery. Additionally, we design the window selective scan module (WSSM), which effectively extracts local haze region information and their differences from other regions, improving the perception of haze distribution and local changes. Extensive experimental results demonstrate that HDMba outperforms state-of-the-art methods in HSI dehazing performance and computational complexity. In future work, we will explore developing weakly supervised and generalized foundation models for HSI dehazing based on the proposed method.

\bibliographystyle{IEEEtran}
\bibliography{ref}
\newpage

\vfill

\end{document}